\documentclass{article}

\PassOptionsToPackage{numbers}{natbib}
\usepackage[final,sglblindworkshop]{neurips_2026}
\workshoptitle{LIGHT: Lightweight Trustworthy Foundation Models}

\usepackage[utf8]{inputenc}
\usepackage[T1]{fontenc}
\usepackage[hidelinks]{hyperref}
\usepackage{url}
\usepackage{booktabs}
\usepackage{amsfonts}
\usepackage{amsmath}
\usepackage{nicefrac}
\usepackage{microtype}
\usepackage{xcolor}
\usepackage{graphicx}
\usepackage{multirow}

\title{When Masking Helps or Hurts Robustness in Compressed CLIP: A Pre-Deployment Diagnostic}

\author{%
  Muhammad Zawish \\
  Technological University Dublin, Ireland \\
  \And
  Steven Davy \\
  Technological University Dublin, Ireland \\
}

\begin{document}

\maketitle

\begin{abstract}
This paper demonstrate that whether masking-based token pruning helps or hurts worst-group robustness can be predicted before deployment, without labels or fine-tuning. A systematic study of semantic masking across 8 spurious-correlation benchmarks shows its effect on worst-group accuracy is highly unstable: it improves accuracy by up to 82.5\% relative on some datasets and degrades it by up to 100\% on others. We trace this instability to spurious inversion: background patches receive higher CLIP text-similarity than the true object when the spurious attribute is background-separable, inverting the assumption every text- and attention-guided pruning method relies on. We introduce the Spurious Inversion Metric (SIM), a label-free, pre-deployment diagnostic whose sign predicts this effect with statistical significance (binomial $p=0.035$) across all 8 datasets, and remains dependable across 6 CLIP architectures with a clean foreground/background split. Naive masking is itself a major source of risk: it causes the largest average-accuracy loss of any method we evaluate, and its own per-image segmentation step is a significant runtime bottleneck. To address this, we design a batched, synchronization-free GPU segmentation routine that cuts this overhead from 3.5$\times$ to 1.75$\times$ baseline. Gating deployment by SIM's sign recovers masking's benefits while avoiding its worst failures, matching or exceeding a strong pruning baseline on 7 of 8 datasets.
\end{abstract}

\section{Introduction}
\label{sec:intro}
\vspace*{-2mm}
CLIP (Contrastive Language-Image Pre-training) \citep{clip} and other contrastively pretrained image-text models now underlie image classification and retrieval systems deployed under real compute budgets. Token pruning is among the most direct routes to compressing them: dropping a fraction of a vision transformer's patch tokens before the transformer body reduces quadratic attention cost, with only a modest drop in clean accuracy. Prior work has approached this through attention-based pruning at an intermediate layer \citep{fastv}, token merging and fusion \citep{evit}, and learned importance scores \citep{pact}, typically evaluated on clean, in-distribution accuracy alone. A parallel line of work targets a different goal: not efficiency, but robustness. The spurious-correlation literature \citep{groupdro} has established that a model's errors are often driven by non-causal background context rather than the object itself, with Waterbirds \citep{groupdro} and CelebA \citep{celeba} as standard benchmarks, later extended by UrbanCars \citep{urbancars}, MetaShifts \citep{metashifts}, and ImageNet-9 \citep{imagenet9}. \emph{Semantic masking}, zeroing out image regions identified as background before encoding, is a natural response: removing the background should also remove the model's reliance on it.

None of the pruning methods above target worst-group robustness, and none of the masking methods above target efficiency. Because masking also reduces the number of tokens that must be processed, it is natural to ask whether a single mechanism can deliver both at once. We call our own construction MARS (Masking And Reduction for Spurious-correlation robustness): it segments each image into foreground and background without supervision, masks out the background pixels, and prunes the remaining tokens. As far as we know, no prior published method combines unsupervised segmentation with pixel-level masking and evaluates it for worst-group robustness on CLIP; MARS is built to test this premise directly, not a reproduction of an existing technique.

Effective use of masking as a compression mechanism entails two core challenges: (1) its effect on worst-group accuracy must be predictable before deployment, since masking can substantially degrade accuracy, and (2) any mechanism used to make this prediction, or to apply masking itself, must be efficient enough that its overhead does not offset the compression it delivers. We show that this combination is not free, and that its failure mode is silent. Across 8 standard spurious-correlation benchmarks, a representative semantic masking method (mask, then prune to half the tokens) improves worst-group accuracy by as much as 82.5\% relative on UrbanCars and 25.6\% on Waterbirds, while degrading it by 70.0\% relative on ImageNet-9 and 39.7\% on MetaShifts. There is no signal at inference time that distinguishes these two regimes: the method is applied identically, and only the eventual worst-group evaluation reveals which outcome occurred. For a paradigm centered on deploying compact models into regulated or safety-relevant settings, an intervention whose sign is unknown in advance is a liability rather than a solution.

We locate the source of this instability in a specific, measurable property of CLIP's joint embedding space: on datasets where the spurious attribute is a separable background region, background patches have \emph{higher} cosine similarity to a generic class-level text prompt than the patches covering the true object, inverting the assumption every text- or attention-guided pruning method relies on (higher text-relevance as evidence for the object). We call this \emph{spurious inversion}: a measurable, dataset-conditional signal, not a property of one benchmark.

Because the underlying quantity (per-patch text similarity, split by an unsupervised foreground/background segmentation) requires no downstream task label and no model evaluation, it can be computed for any candidate deployment dataset before a single classification is scored. We formalize this as the Spurious Inversion Metric (SIM): a diagnostic, not a corrective mechanism, that tells a practitioner in advance whether masking is likely to help or hurt on their data. Section~\ref{sec:method} makes this precise: our implementation samples images according to the same group structure (label $\times$ spurious attribute) worst-group evaluation itself uses, so it presumes that structure is known, though it needs neither correctness labels nor a trained classifier.

\textbf{Contributions.} (1) We document spurious inversion, a measurable, mechanistically grounded property of CLIP representations, distinct from the known fact that attention alone is a biased importance signal. (2) We introduce SIM, a diagnostic requiring no downstream task label and no model evaluation, whose sign predicts the direction of masking's effect on worst-group accuracy with statistical significance across 8 datasets (binomial $p=0.035$; Spearman $\rho=0.78$). (3) We show gating deployment by SIM's sign matches or exceeds blind masking on every dataset, and a strong fixed pruning baseline on all but the one dataset where SIM's blind spot misfires, recovering most of masking's benefit without most of its worst failures. (4) We extend the analysis to 6 CLIP variants and characterize where the diagnostic holds and where it does not: dependable with a clean foreground/background separation, with a specific, reproducible blind spot on multi-context spurious structure (MetaShifts), which we report rather than obscure. (5) We separate two costs prior framings of masking-based compression conflate: SIM is a dataset-level diagnostic amortized to zero at inference time, while MARS's own segmentation step is a genuine per-inference cost; we document two unsuccessful attempts to remove this cost and one that succeeds, reducing MARS's overhead from 3.5$\times$ to roughly 1.75$\times$ baseline and leaving the remaining gap as an explicit open problem.

\begin{figure}[t]
  \centering
  \includegraphics[width=0.82\linewidth]{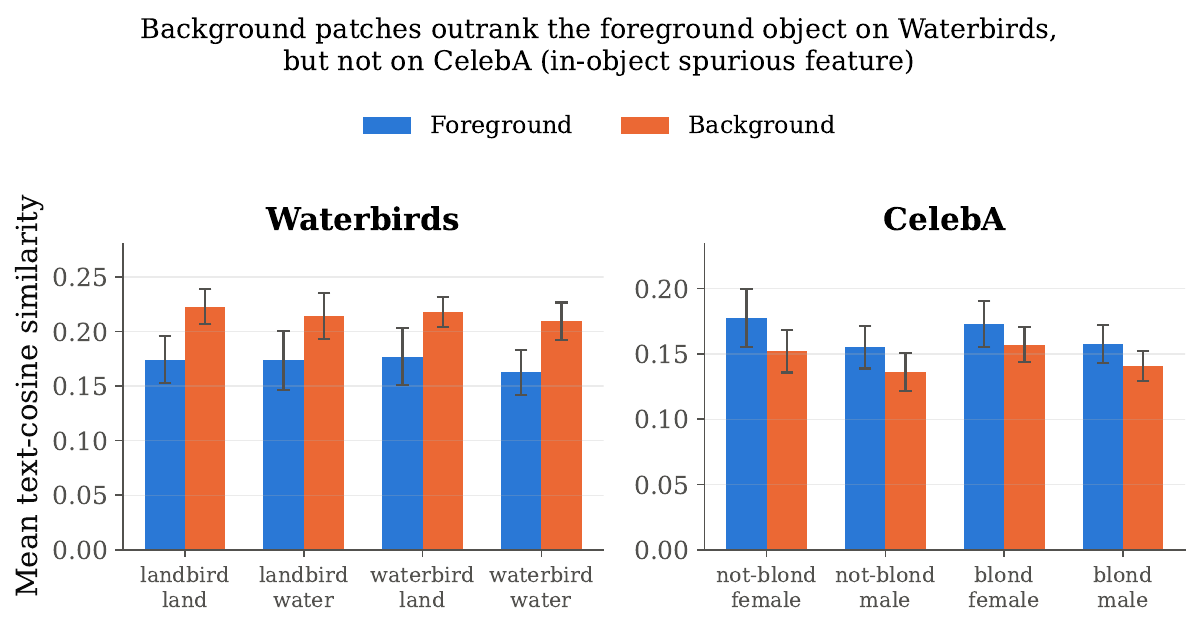}
  \caption{Foreground vs.\ background text-similarity on Waterbirds (background-spurious) and CelebA (in-object-spurious). Background patches outrank the object on Waterbirds across every group; the reverse holds on CelebA. Error bars: one standard deviation across images within each group.}
  \label{fig:inversion}
\end{figure}
\vspace*{-3mm}
\section{Related Work}
\label{sec:related}
\vspace*{-3mm}
\textbf{Token pruning for efficient vision transformers.} FastV \citep{fastv} prunes tokens by attention magnitude at an intermediate layer, EViT \citep{evit} reorganizes and fuses low-attention tokens into the class token, and PACT \citep{pact} combines attention-based pruning with clustering-based merging, all evaluated on clean or downstream accuracy, not worst-group robustness. ATP \citep{atp} reports attention-guided pruning can suppress spurious features under visual corruption, though this observation is preliminary and dataset-specific. Our compression-risk analysis (Section~\ref{sec:results-risk}) is, to our knowledge, the first systematic evaluation of pruning's effect on worst-group robustness across many datasets and methods at once.

\textbf{Spurious correlations and worst-group robustness.} Waterbirds \citep{groupdro} and CelebA \citep{celeba}, the latter adopted for group-robustness evaluation with the group splits from \citep{groupdro}, established worst-group accuracy as the standard metric for spurious-attribute reliance, later extended by UrbanCars \citep{urbancars}, MetaShifts \citep{metashifts}, and background-manipulation benchmarks such as ImageNet-9 \citep{imagenet9} and ImageNet-D \citep{imagenetd}; a recent survey \citep{cleverhans} situates these within the broader landscape of spurious correlations across machine learning. A separate line of work addresses spurious correlations directly through masking or attention-reweighting informed by unsupervised segmentation; our contribution is complementary: rather than proposing a new masking mechanism, we characterize when an existing one should or should not be trusted, using a diagnostic that needs no downstream task label and no additional model evaluation beyond what the segmentation step already computes.

\textbf{CLIP robustness diagnostics.} Prior work has noted that CLIP's attention is biased toward image regions correlated with, but not necessarily causally responsible for, the predicted label. Recent work corrects this bias directly, either by searching over prompts to reduce reliance on spurious multimodal cues \citep{sage} or by locating and editing the specific attention heads responsible for it \citep{debiasclip}. Our finding is a distinct and more specific claim: it is not attention alone that is biased, but the text-similarity signal itself that inverts, in a way that is measurable per-dataset without labels and predicts a specific downstream failure mode.

\section{Spurious Inversion and the SIM Diagnostic}
\label{sec:method}

\textbf{Setup.} We build on OpenCLIP ViT-L/14 (\texttt{laion2b\_s32b\_b82k}), which tokenizes a $224\times224$ image into a $16\times16$ grid of 256 patch tokens. For a given image, we compute a foreground/background partition using PCA (3 components) on the patch embeddings, a Gaussian smoothing pass over the resulting spatial grid ($\sigma=0.72$), and $k$-means ($k=3$) clustering. The cluster whose mean embedding has the lower cosine similarity to a generic class-level text prompt (e.g., ``a photo of a bird'') is treated as the object segment on datasets where the spurious attribute is expected to be background; the convention is reversed for datasets, such as CelebA, where the spurious attribute is itself part of the foreground. \emph{MARS} refers to the masking-and-pruning method we study throughout: pixels outside the selected foreground segment are zeroed, and the resulting image is re-encoded with only the top 50\% of patch tokens (ranked by foreground-cluster membership) retained.

\begin{figure}[t]
  \centering
  \includegraphics[width=0.9\linewidth]{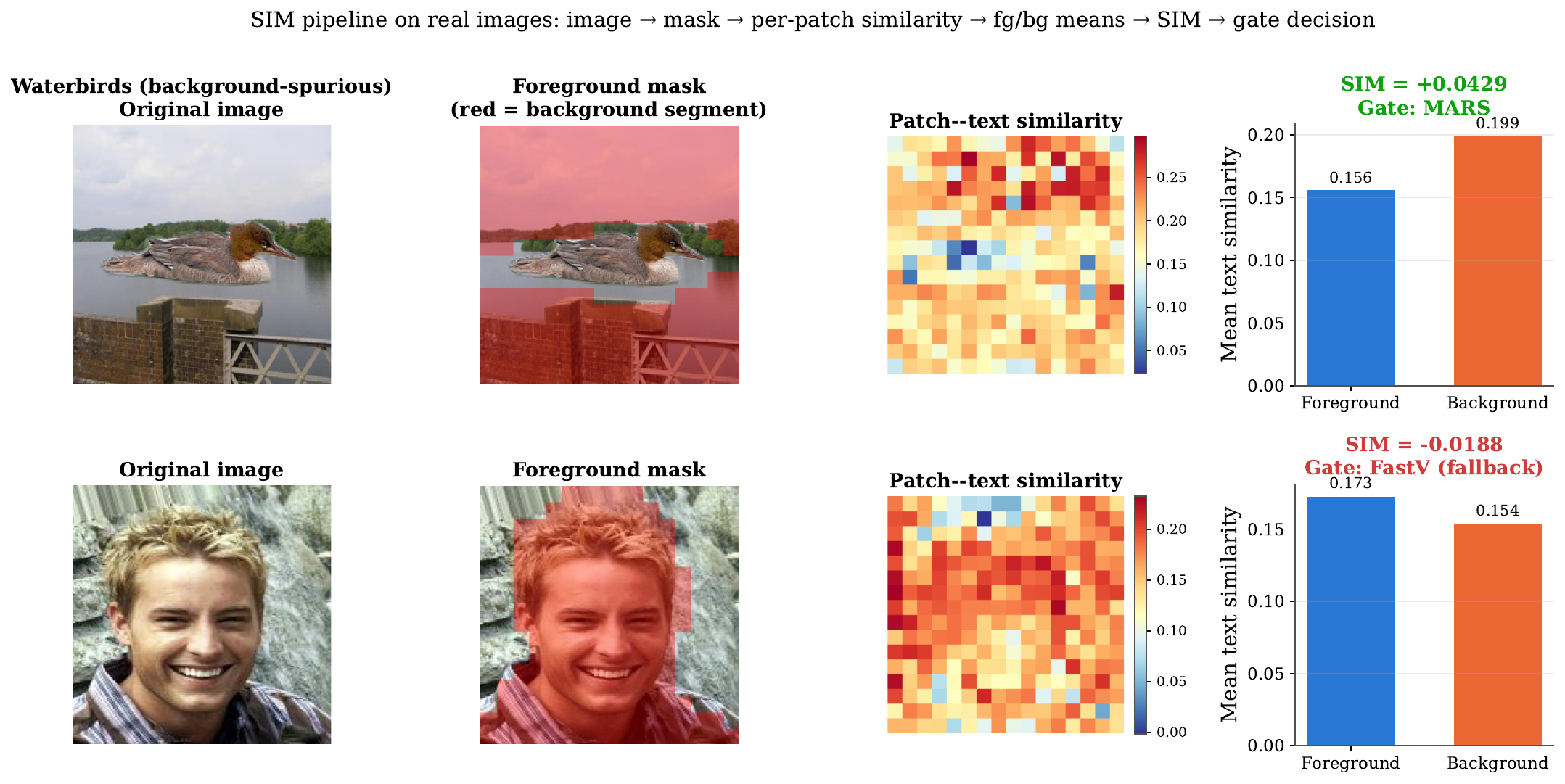}
  \caption{The SIM pipeline on two real, representative images (selected as the image within each dataset closest to that dataset's own mean SIM, not a cherry-picked extreme): original image, the unsupervised foreground/background mask, the per-patch text-similarity heatmap, and the resulting foreground/background means, SIM value, and gate decision. On Waterbirds, background patches (sky, water, wall) outrank the bird itself, giving positive SIM and a MARS decision; on CelebA, the foreground face outranks its surroundings, giving negative SIM and a fallback-to-FastV decision.}
  \label{fig:pipeline}
\end{figure}

\textbf{Spurious inversion.} For each dataset and each foreground/background split, we measure the mean cosine similarity between the general text prompt and (i) the patches assigned to the foreground segment and (ii) the patches assigned to the background segment. On Waterbirds and UrbanCars, background patches have consistently \emph{higher} text-similarity than foreground patches, across every group (Figure~\ref{fig:inversion}, left). We attribute this to contrastive pretraining on web image-caption pairs, where a background co-occurring reliably with a class label (water for waterbirds, urban scenery for city-registered cars) inflates that background's alignment with the class prompt in the joint embedding space; the next paragraph makes this account precise. On CelebA, where the spurious attribute (gender) is encoded in the foreground face rather than the background, no such inversion occurs (Figure~\ref{fig:inversion}, right): foreground similarity exceeds background similarity, as expected under a non-inverted regime. Figure~\ref{fig:pipeline} walks through this computation end-to-end on two representative images.

\textbf{Why inversion occurs where it does.} A contrastively trained joint embedding aligns an aggregate image representation with its class-level text prompt across the training distribution, and that aggregate depends on both object and background regions. When an object category has substantial visual variation (bird species, body pose, car models) while a co-occurring background is comparatively stable, the background becomes the lower-variance, more consistently label-aligned signal, even though the label is semantically about the object. Background patches can therefore end up more text-aligned than object patches when the spurious attribute is background and the object is heterogeneous, the condition met by Waterbirds and UrbanCars but not CelebA, where the class-relevant signal (gender) sits in a stable facial region. This also predicts the boundary case in Section~\ref{sec:results-arch}: MetaShifts spreads its spurious attribute over 11 distinct contexts, so no single context is stable enough to reliably outrank the object, a plausible reason SIM's sign is unreliable there.

\textbf{The Spurious Inversion Metric.} We define, for a dataset $D$ with groups $g \in \{1, \dots, G\}$,
\begin{equation}
  \mathrm{SIM}(D) = \frac{1}{G}\sum_{g=1}^{G} \left(\bar{s}_{\text{bg}}(g) - \bar{s}_{\text{fg}}(g)\right),
\end{equation}
where $\bar{s}_{\text{fg}}(g)$ and $\bar{s}_{\text{bg}}(g)$ are the mean patch-to-text cosine similarities of the foreground and background segments, averaged over sampled images in group $g$. SIM needs no ground-truth segmentation (the split comes from the same unsupervised PCA-and-$k$-means procedure MARS itself uses) and no downstream task label or model evaluation: unlike worst-group accuracy, SIM never queries the classifier's output. It does require, in the form evaluated here, a modest sample stratified across the dataset's existing groups $g \in \{1, \dots, G\}$ (label $\times$ spurious attribute), the same structure worst-group accuracy is defined over, meaningfully weaker than a labeled evaluation (no correctness labels, no trained classifier, one dot product per patch instead of a full evaluation pass). We report our main results using this group-stratified estimate for exact comparability with the metric it predicts; Appendix~\ref{app:nonstratified} confirms a non-stratified estimate recovers the same sign-prediction performance. Our central claim: $\mathrm{sign}(\mathrm{SIM}(D))$ predicts whether applying MARS to $D$ will increase or decrease worst-group accuracy relative to the uncompressed baseline.

\textbf{A gated deployment policy.} SIM's practical use is as a pre-deployment gate: apply MARS if $\mathrm{SIM}(D) > 0$; otherwise, fall back to a fixed, non-masking pruning baseline at the same token budget. This policy never processes more than half the tokens a full-resolution forward pass would use, regardless of which branch is taken; the only decision is which compression mechanism to trust for a given deployment target.

We deliberately choose the simplest policy the data supports. The fallback method is FastV, not an arbitrary choice: Table~\ref{tab:risk} shows FastV is tied with FiCoCo for the best mean change in worst-group accuracy of any pruning method we evaluate, and markedly faster (12.4ms vs.\ 15.7ms per image), making it the appropriate default whenever a dataset's structure does not support masking. The gate itself is a bare sign test, with no fitted threshold or learned classifier: our evidence base is only 8 datasets, and a tuned threshold would overfit a sample this size. Section~\ref{sec:results-arch} confirms this: a tuned threshold, chosen to correct one dataset's misclassification, transfers worse across architectures than the parameter-free sign test it was meant to improve on. A learned gate remains a natural extension once more labeled deployment datasets are available.
\vspace*{-2mm}
\section{Experimental Setup}
\label{sec:setup}
\vspace*{-2mm}
\textbf{Datasets.} We evaluate on 8 datasets spanning three structural regimes. Waterbirds \citep{groupdro}, UrbanCars \citep{urbancars}, and MetaShifts \citep{metashifts} have a stable, naturalistically co-occurring background-class correlation (worst-group accuracy over label $\times$ spurious-attribute groups), where MARS is expected to help. CelebA \citep{celeba} has an in-object spurious attribute (gender, in the foreground rather than the background), where masking is expected to hurt. The remaining four lack an exploitable background-class correlation, for different reasons: ImageNet-9 \citep{imagenet9} is background-manipulated, but its group split (original vs.\ backgrounds swapped in at random from other classes) has no stable co-occurrence for CLIP's pretraining to have learned, unlike Waterbirds' or UrbanCars' naturalistic pairings; OxfordPets \citep{oxfordpets}, Caltech101 \citep{caltech101}, and ImageNet-D \citep{imagenetd} define worst-group accuracy over class identity, not a background split (ImageNet-D's images are uniformly background-shifted, so no clean-background group exists to compare against). Masking is expected to hurt or be neutral on all four.

\textbf{Models.} Our primary study uses OpenCLIP ViT-L/14 (\texttt{laion2b\_s32b\_b82k}). Section~\ref{sec:results-arch} extends this to 6 CLIP variants spanning two independent axes: model size (ViT-B/32, ViT-B/16, ViT-L/14, ViT-H/14) and training data (OpenAI's original CLIP corpus vs.\ LAION-2B, held constant in architecture at both ViT-B/32 and ViT-L/14).

\textbf{Baselines.} We compare against five standard token pruning methods that require no masking: FastV \citep{fastv} (attention-based pruning at an intermediate layer), EViT \citep{evit} (attention pruning with token fusion), PatchRank \citep{patchrank} (multi-layer attention-norm ranking), PACT \citep{pact} (attention pruning with token merging), and FiCoCo \citep{ficoco} (redundancy-based token filtering), together with a text-guided pruning baseline (ranking tokens by direct text similarity, in the style of SparseVLM \citep{sparsevlm}) that we include specifically because it is the pruning strategy spurious inversion most directly undermines. All pruning baselines and MARS operate at a 50\% token budget (128 of 256 tokens) unless noted otherwise.

\textbf{Statistical reporting.} We report mean $\pm$ standard deviation over 3 random seeds for Waterbirds, UrbanCars, and CelebA (full breakdown in Appendix~\ref{app:seeded}). For UrbanCars and CelebA, whose test splits are smaller than the sampled evaluation size, this variation reflects segmentation-algorithm sensitivity rather than resampling variance, so the two are not on the same statistical footing as the rest of the table. All other datasets are evaluated on a single full pass given their larger test splits.

\section{Results}
\label{sec:results}

\subsection{Compression risk varies by dataset}
\label{sec:results-risk}

Table~\ref{tab:risk} reports each method's mean relative change in worst-group accuracy vs.\ the uncompressed baseline, its average accuracy, and its per-image latency, averaged over 8 datasets. Five of six standard baselines carry only mild worst-group risk (within about 7\% relative); FastV and FiCoCo mildly \emph{improve} it on average. Two methods are catastrophic outliers: text-guided pruning (36.7\% mean relative loss) and MARS applied without gating (31.6\%). Text-guided pruning is riskiest on worst-group accuracy; MARS has the lowest average accuracy of any method: 48.2\% vs.\ 79.9\% for the uncompressed baseline, a 38.3\% mean per-dataset relative drop (not $(79.9-48.2)/79.9=39.7\%$, which applies that formula to the two aggregates directly), the largest drop we observe, showing unconditional masking damages overall usability, not just worst-group robustness. Latency also varies more than attention alone would predict: PACT's greedy token-merging step is the slowest single-pass method we benchmark, and text-guided pruning's hidden two-pass structure (score all 256 tokens, then re-encode from scratch with 128) makes it markedly slower than FastV, EViT, PatchRank, and FiCoCo, though still faster than PACT and MARS. MARS remains the slowest method overall (Appendix~\ref{app:latency}). Unconditionally deploying MARS or text-guided pruning is measurably riskier than deploying no masking-aware compression at all.

\begin{table}[t]
  \caption{Mean relative change in worst-group accuracy vs.\ the uncompressed baseline (\%), average accuracy (\%), and per-image latency (batch size 1; MARS latency uses the original CPU segmentation step, Appendix~\ref{app:latency}), averaged over 8 datasets, for each pruning method at a 50\% token budget. Hurt/Helped count datasets with strictly negative/positive $\Delta$WG; the remainder show negligible ($\approx$0) change.}
  \label{tab:risk}
  \centering
  \small
  \setlength{\tabcolsep}{4.5pt}
  \begin{tabular}{lccccc}
    \toprule
    Method & Rel.\ $\Delta$WG (\%) & Avg.\ Acc.\ (\%) & Latency (ms) & Hurt & Helped \\
    \midrule
    Baseline (uncompressed) & - & 79.9 & 15.7 & - & - \\
    FastV \citep{fastv}           & $+2.3\%$  & 79.7 & 12.4 & 2/8 & 3/8 \\
    FiCoCo \citep{ficoco}          & $+2.3\%$  & 78.9 & 15.7 & 4/8 & 3/8 \\
    EViT \citep{evit}             & $-4.3\%$  & 79.7 & 12.7 & 4/8 & 2/8 \\
    PatchRank \citep{patchrank}    & $-5.2\%$  & 80.8 & 15.1 & 4/8 & 2/8 \\
    PACT \citep{pact}             & $-7.1\%$  & 78.5 & 34.1 & 5/8 & 2/8 \\
    MARS (blind, ours) & $-31.6\%$ & 48.2 & 55.2 & 5/8 & 2/8 \\
    Text-guided \citep{sparsevlm} & $-36.7\%$ & 59.1 & 26.6 & 7/8 & 0/8 \\
    \bottomrule
  \end{tabular}
\end{table}
\vspace*{-3mm}
\subsection{SIM predicts masking's effect}
\label{sec:results-sim}

\begin{figure}[t]
  \centering
  \includegraphics[width=0.5\linewidth]{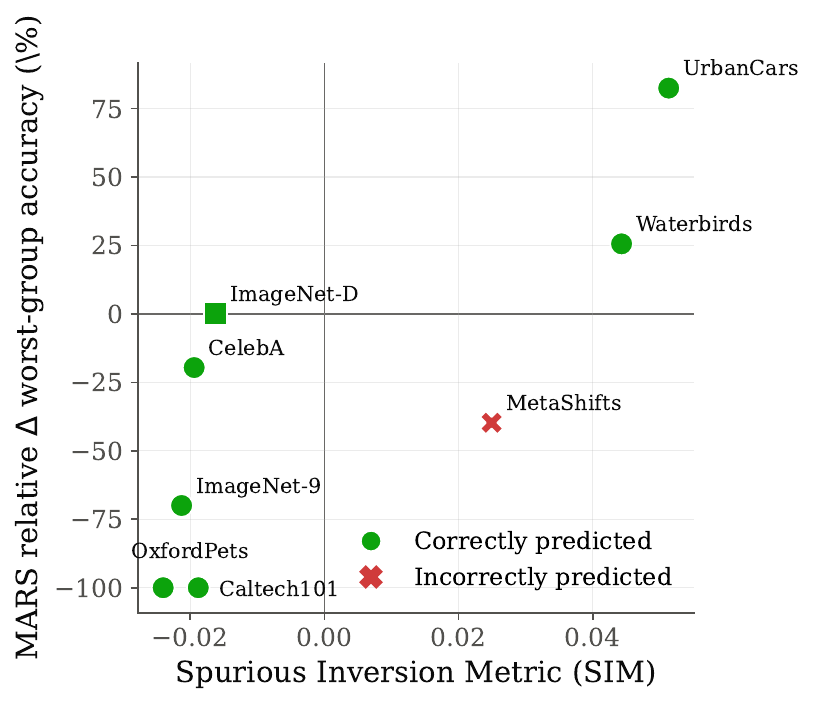}
  \caption{SIM against MARS's measured change in worst-group accuracy. Seven of eight datasets fall on the side of zero that SIM predicts; the single exception, MetaShifts, is analyzed as a structural boundary case in Section~\ref{sec:results-arch}.}
  \label{fig:sim}
\end{figure}

Figure~\ref{fig:sim} plots SIM against MARS's actual measured change in worst-group accuracy for all 8 datasets (full per-dataset values in Table~\ref{tab:sim} of Appendix~\ref{app:sim-full}). SIM's sign correctly predicts the direction of the effect on 7 of 8 datasets. Since this is fundamentally a binary classification of direction rather than a linear relationship, we report an exact one-sided binomial test against a 50\% chance baseline: $p=0.035$. This scores ImageNet-D, whose baseline and MARS worst-group accuracy are both exactly 0.0\%, as a correct ``hurt'' prediction, since SIM's sign is negative there and a non-positive difference is the same directional outcome as a negative one; excluding this degenerate tie entirely gives 6 of 7 datasets and $p=0.063$, not significant at the conventional 0.05 threshold. We report the inclusive figure as primary but flag this sensitivity explicitly; Section~\ref{sec:limitations} returns to it. The Spearman rank correlation between SIM and MARS's relative change gives $\rho=0.78$ ($p<0.05$, $t=2.74$ on 5 degrees of freedom, $n=7$; ImageNet-D excluded here too, since its relative change is undefined). The one miss, MetaShifts, is analyzed in Section~\ref{sec:results-arch} and Section~\ref{sec:limitations}, where we show it is a reproducible, structural property of that dataset rather than an isolated artifact.

SIM's magnitude, not only its sign, carries information about how much a prediction should be trusted. MetaShifts has the smallest magnitude among the three datasets where SIM is positive ($+0.025$, against $+0.044$ and $+0.051$ for Waterbirds and UrbanCars), and this smallest-magnitude case is precisely the one that produces the incorrect sign prediction, both here and, as we show next, across every architecture we test. A value of SIM close to zero marks an implicit low-confidence region for the sign test, much like a classifier's decision boundary. A natural extension is a policy that abstains, or defaults to the safer fallback, when $|\mathrm{SIM}(D)|$ falls below a data-driven margin; constructing and validating such a margin would need a larger collection of datasets than the 8 available to us, and we leave it as a concrete next step.

\subsection{Gating recovers masking's benefit}
\label{sec:results-gate}

We evaluate the gated policy from Section~\ref{sec:method}: apply MARS when $\mathrm{SIM}(D) > 0$, otherwise FastV, at a fixed 128-token budget. Table~\ref{tab:gate} compares this against always applying MARS (blind) and always applying FastV (blind) across all 8 datasets (Figure~\ref{fig:gate}, Appendix~\ref{app:gate-fig}). The gated policy matches or exceeds blind MARS on every dataset by construction, and matches FastV exactly whenever it correctly abstains from masking, 7 of 8 datasets. The one exception: on MetaShifts, the gate follows SIM's incorrect positive sign and selects MARS, scoring 52.2\% against the 87.8\% blind FastV would have achieved, a 40.5\% relative deficit: the direct practical cost of the diagnostic's one structural blind spot (Section~\ref{sec:results-sim}), and why the gate's mean (49.4\%) sits slightly below blind FastV's mean (51.1\%). The gate strictly improves only over always running MARS blindly, not over FastV; Section~\ref{sec:limitations} treats this as a central finding, not a footnote.

The right way to read this comparison is bounding risk under uncertainty about the deployment target, not maximizing mean accuracy. A practitioner who does not know whether a new dataset resembles Waterbirds and UrbanCars, where blind FastV forfeits 11.9\% and 35.0\% of MARS's worst-group accuracy, or MetaShifts, where blind MARS risks a 39.7\% relative loss, faces a genuine bet either way. The gate removes the need to make that bet blindly: its known downside is a 40.5\% relative loss on one dataset out of eight, bounded by the structural condition in Section~\ref{sec:results-arch}, not unbounded exposure to whichever failure mode a new target triggers.

This evaluation is a sign test with no free parameter: no threshold was tuned on this table. An earlier version used a small positive margin (threshold $0.03$) to additionally correct the MetaShifts case, but Section~\ref{sec:results-arch} shows this margin does not transfer across architectures as reliably as the parameter-free sign test. We therefore report the sign-test result, the version of the claim we can defend across the full range of models evaluated.

\begin{table}[t]
  \caption{Worst-group accuracy (\%) under three deployment policies, all at 128 tokens except the uncompressed baseline (256 tokens).}
  \label{tab:gate}
  \centering
  \small
  \setlength{\tabcolsep}{4.5pt}
  \begin{tabular}{lrrrr}
    \toprule
    Dataset & Baseline (256 tok) & Blind MARS & Blind FastV & SIM-gated \\
    \midrule
    Waterbirds \citep{groupdro}  & 50.8 & \textbf{67.3} & 59.3 & \textbf{67.3} \\
    UrbanCars \citep{urbancars}  & 22.8 & \textbf{40.3} & 26.2 & \textbf{40.3} \\
    CelebA \citep{celeba}     & 79.6 & 64.7 & \textbf{75.6} & \textbf{75.6} \\
    MetaShifts \citep{metashifts} & 86.7 & 52.2 & \textbf{87.8} & 52.2 \\
    ImageNet-9 \citep{imagenet9} & 87.5 & 26.3 & \textbf{85.3} & \textbf{85.3} \\
    OxfordPets \citep{oxfordpets} & 52.6 & 0.0  & \textbf{52.6} & \textbf{52.6} \\
    Caltech101 \citep{caltech101} & 22.2 & 0.0  & \textbf{22.2} & \textbf{22.2} \\
    ImageNet-D \citep{imagenetd} & 0.0  & 0.0  & 0.0  & 0.0  \\
    \midrule
    Mean        & 50.3 & 31.4 & 51.1 & \textbf{49.4} \\
    \bottomrule
  \end{tabular}
\end{table}

\subsection{Generalization across architectures}
\label{sec:results-arch}

A diagnostic validated on a single checkpoint is of limited practical use. We extend the sign-test evaluation to 6 CLIP variants: ViT-B/32 and ViT-L/14 each under both OpenAI and LAION-2B pretraining (isolating the training-data axis at matched architecture), ViT-B/16, and ViT-H/14 (extending the capacity axis), evaluated across all 8 datasets, for 48 (architecture, dataset) combinations in total.

The sign test is correct on 31 of 48 combinations (64.6\%; exact binomial $p=0.030$). These 48 trials are not independent: they are 8 datasets repeated across 6 architectures, and a dataset's outcome is correlated across architectures (MetaShifts contributes 6 dependent failures, not 6 independent ones), so the effective sample size is closer to 8 than 48. Figure~\ref{fig:heatmap} (Appendix~\ref{app:arch-full}) visualizes the pattern, concentrated rather than noisy. Five of the six architectures are correct on all three datasets the diagnostic was designed around (Waterbirds, UrbanCars, CelebA); the sixth (ViT-H/14, LAION) misses only Waterbirds. MetaShifts is incorrect for all 6 architectures without exception, an architecture-independent property of its subtler, multi-context spurious attribute (11 scene contexts, not one background category). A smaller number of misses occur on the fine-grained benchmarks (OxfordPets, Caltech101, ImageNet-D), several coinciding with both methods collapsing to near-zero accuracy regardless of the gate's decision, a separate phenomenon from spurious inversion.

Notably, ViT-L/14 under LAION-2B and OpenAI pretraining achieve closely comparable correctness at matched architecture: 6/8 and 5/8, respectively, differing only on Caltech101, a fine-grained benchmark already identified above as a separate failure mode. This indicates that spurious inversion is a property of contrastive image-text pretraining broadly, not an artifact specific to the LAION-2B corpus. We also verify that the margin used in Table~\ref{tab:gate}'s development (threshold $0.03$) does not outperform the parameter-free sign test once evaluated across architectures: both achieve exactly 31/48. This is the reason we report the sign test, not a tuned threshold, as our primary claim throughout this paper.

\subsection{Two distinct costs: offline vs.\ per-inference}
\label{sec:results-efficiency}
\vspace*{-3mm}
A compression pipeline built around SIM incurs two costs at different points in the deployment lifecycle. \emph{SIM is a dataset-level quantity}: computing it requires sampling a modest number of images from the target deployment distribution once, before any inference-time decision is made, so its cost is a fixed, one-time overhead amortized to zero over the deployment's lifetime. \emph{MARS's own masking step is different}: whenever the gate selects MARS, every subsequent image still requires its own foreground/background segmentation, since the mask is specific to that image's content, not to the dataset as a whole.

Full per-stage latency is reported in Table~\ref{tab:latency} of Appendix~\ref{app:latency}. One detail matters for reading MARS's cost correctly: the patch embeddings that feed PCA and $k$-means must be the \emph{contextualized} output of the full vision transformer (Section~\ref{sec:method}), requiring one full, unpruned 256-token forward pass before segmentation can begin; MARS's masked, pruned re-encoding at 128 tokens is a second forward pass on top of that. Summing all three stages, MARS's true per-image cost is $\approx$15.7ms + 28.5ms (CPU segmentation) + 11.0ms $\approx$55.2ms, roughly 3.5$\times$ the 15.7ms baseline. FastV incurs no such first pass, since it prunes from attention weights already computed inside its single forward pass, for a complete one-pass cost of 12.4ms.

We investigated three independent routes to reducing this segmentation cost; two did not succeed. Appendix~\ref{app:negative} reports both in full: a single-image GPU reimplementation, which was slower than the CPU path it targeted (161ms/image vs.\ 28.5ms) and less faithful to it (65\% mask IoU), and an attention-based diagnostic that showed no dataset-discriminative signal at all.

The third route succeeded. Profiling the single-image GPU attempt's $k$-means step revealed the real problem: its initialization and convergence check called \texttt{.item()} and \texttt{.cpu()} every iteration, a host-device synchronization that idles the GPU once per image. We removed every such call and batched PCA-and-$k$-means across many images in one vectorized call. This reduces segmentation cost to 0.81ms/image at batch size 128, a $35\times$ reduction from the CPU path. We validated this before trusting it: on 300 held-out images each from Waterbirds, UrbanCars, and CelebA, its worst-group accuracy matched or exceeded the CPU reference on all three (full sweep in Appendix~\ref{app:negative}). With this fix, MARS's true per-image cost falls to $\approx$27.5ms, roughly 1.75$\times$ baseline, about half the original 3.5$\times$ gap. We also tried reusing one fixed PCA basis across a dataset rather than refitting it per image; this is faster still (0.14ms/image) but less accurate and consistent ($-9.4\%$ to $+21.8\%$ relative, vs.\ the per-image variant's $0.0\%$ to $+6.5\%$), so we report the per-image-adaptive result as our validated finding.
\vspace*{-3mm}
\section{Discussion and Limitations}
\label{sec:limitations}
\vspace*{-2mm}
\textbf{Where the diagnostic is reliable.} SIM's sign is dependable on datasets whose spurious attribute is a spatially coherent, separable region: background scenery, in our benchmarks. Section~\ref{sec:results-arch} shows near-perfect sign-agreement on the three datasets with this property (Waterbirds, UrbanCars, CelebA: 17 of 18 correct across 6 architectures), in sharp contrast to its one systematic failure (MetaShifts, incorrect for all 6 architectures), whose spurious attribute (11 distinct scene contexts) does not fit this description. This boundary is a contribution in its own right: it names the condition under which the diagnostic can be expected to work.

\textbf{Group-stratified sampling is not essential.} Our main results compute SIM from samples stratified across each benchmark's existing groups, a weaker requirement than the labeled evaluation it replaces, but not the same as a fully unlabeled pool. A flat, unstratified pool of 200 images per dataset, with no group indexing at all (Appendix~\ref{app:nonstratified}), achieves the same 7 of 8 correct sign predictions, agreeing with the stratified version on all 8 datasets. This is reasonably strong, though not exhaustive, evidence that SIM's practical requirement is a representative unlabeled sample, not knowledge of group structure.

\textbf{The sign test rests on 8 datasets.} The headline binomial result (7 of 8 correct, $p=0.035$) is not robust to the small sample it is computed over, and this is not merely a hypothetical concern about future data: Section~\ref{sec:results-sim} shows it already hinges on how ImageNet-D's degenerate 0-vs-0 tie is scored, with the excluded-tie alternative (6 of 7, $p=0.063$) crossing the conventional significance threshold. We corroborate the inclusive result with the architecture-generalization sweep (Section~\ref{sec:results-arch}) and the Spearman correlation, both pointing the same direction, but none of these substitutes for a larger and more diverse collection of benchmark datasets than the field currently offers for this problem.

\textbf{Validation is retrospective.} Every result here checks SIM's prediction against a worst-group accuracy measurement we already had. A genuinely prospective test, committing to a prediction before any labeled evaluation exists and then measuring the outcome, is the appropriate next step; we attempted this using two additional benchmarks (Spawrious and CounterAnimal), never used to develop SIM, but neither was available in a usable state on our compute infrastructure in time for this submission.

\textbf{Efficiency: improved, not resolved.} Section~\ref{sec:results-efficiency} found MARS's segmentation cost initially brought its total latency to roughly 3.5$\times$ baseline; a batched, synchronization-free GPU reimplementation reduces this to roughly 1.75$\times$, validated for worst-group-accuracy parity across three datasets (Appendix~\ref{app:negative}). The remaining gap follows from MARS's design, not an inherent limit of masking-based compression: it masks pixels and re-encodes from scratch, needing two full forward passes regardless of segmentation speed, whereas FastV and EViT prune from an intermediate layer's activations and continue the same pass with fewer tokens. A redesign that segments from intermediate-layer activations within a single pass, rather than re-encoding from the input, is untested here and would plausibly close most of the gap; we leave it to future work.

\textbf{Broader impacts.} This work's main contribution, a diagnostic that flags in advance when masking-based compression will hurt worst-group robustness, is intended to reduce the risk of silently deploying a compressed CLIP-style model that performs worse for a minority subgroup, a failure mode with direct fairness implications. We also see a specific misuse risk: SIM has a documented blind spot (MetaShifts) and is validated on only 8 datasets, so treating a positive SIM-gated decision as a general safety certificate, rather than evidence bounded by the conditions we identify, would misuse it. All datasets used are existing, standard academic benchmarks; we introduce no new datasets, pretrained models, or public releases carrying an independent misuse risk.
\vspace*{-3mm}
\section{Conclusion}
\label{sec:conclusion}
\vspace*{-2mm}
We show that a widely plausible assumption, that masking spurious background context improves both the robustness and the efficiency of a compressed CLIP-style model, holds only conditionally, and that its failure mode is invisible without a diagnostic. Spurious inversion gives a mechanistic account of why standard importance signals point the wrong way on certain datasets, and SIM turns this account into a measurement a practitioner can take before deployment, with a statistically supported and architecture-tested claim about when it can be trusted. We view the precise characterization of where this diagnostic holds, together with a clear accounting of what remains computationally unsolved, as the appropriate empirical contribution of a paper whose subject is when to trust a compression method, not merely whether one particular method is best.

\section*{Acknowledgments}
This publication has emanated from research conducted with the financial support of Taighde \'Eireann -- Research Ireland under Grant number 21/FFP-A/9174.

\bibliographystyle{plainnat}

\clearpage
\appendix

\section{Full 3-seed main table}
\label{app:seeded}

Table~\ref{tab:seeded} reports the complete per-method, per-dataset breakdown underlying the 3-seed statistical claims in Section~\ref{sec:setup}. As noted there, UrbanCars' and CelebA's full test splits are smaller than or equal to our sampled evaluation size, so every seed draws an identical sample; the reported variation for these two datasets reflects the $k$-means segmentation step's own sensitivity to its random initialization, not resampling variance. Waterbirds' $\sim$5,794-image test set is genuinely subsampled differently per seed, so its variation reflects both sources jointly. The three ``3-seed'' columns therefore do not reflect a uniform standard of statistical rigor across datasets, and should not be read as if they did.

\begin{table}[h]
  \caption{Worst-group and average accuracy (\%, mean $\pm$ std over 3 seeds) for all evaluated methods on the three datasets with seeded evaluation.}
  \label{tab:seeded}
  \centering
  \small
  \begin{tabular}{llrrr}
    \toprule
    Dataset & Method & WG acc.\ (\%) & Avg.\ acc.\ (\%) & Tokens \\
    \midrule
    \multirow{11}{*}{Waterbirds \citep{groupdro}} & Baseline & $50.8 \pm 2.4$ & $77.8 \pm 1.0$ & 256 \\
 & Masked (no prune) & $66.5 \pm 3.3$ & $81.0 \pm 0.7$ & 256 \\
 & Semantic-prune-only & $66.3 \pm 2.9$ & $80.1 \pm 0.7$ & 128 \\
 & MARS (hybrid) & $67.3 \pm 2.7$ & $80.3 \pm 0.7$ & 128 \\
 & Text-guided \citep{sparsevlm} & $18.5 \pm 0.8$ & $56.5 \pm 0.3$ & 128 \\
 & FastV \citep{fastv} & $59.3 \pm 3.1$ & $79.4 \pm 1.1$ & 128 \\
 & EViT \citep{evit} & $58.7 \pm 2.7$ & $79.9 \pm 1.0$ & 128 \\
 & PatchRank \citep{patchrank} & $62.7 \pm 0.9$ & $80.6 \pm 0.6$ & 128 \\
 & PACT \citep{pact} & $57.2 \pm 1.8$ & $76.9 \pm 0.8$ & 128 \\
 & FiCoCo \citep{ficoco} & $56.7 \pm 2.7$ & $78.5 \pm 1.0$ & 128 \\
 & MARS (CLS-attn.\ variant) & $57.7 \pm 1.8$ & $78.5 \pm 0.7$ & 128 \\
\midrule
\multirow{11}{*}{UrbanCars \citep{urbancars}} & Baseline & $22.8 \pm 0.0$ & $51.2 \pm 0.0$ & 256 \\
 & Masked (no prune) & $39.3 \pm 0.6$ & $52.6 \pm 0.3$ & 256 \\
 & Semantic-prune-only & $29.0 \pm 0.9$ & $52.7 \pm 0.5$ & 128 \\
 & MARS (hybrid) & $40.3 \pm 1.0$ & $54.6 \pm 0.9$ & 128 \\
 & Text-guided \citep{sparsevlm} & $20.0 \pm 0.0$ & $49.0 \pm 0.0$ & 128 \\
 & FastV \citep{fastv} & $26.2 \pm 0.0$ & $53.1 \pm 0.0$ & 128 \\
 & EViT \citep{evit} & $26.2 \pm 0.0$ & $53.1 \pm 0.0$ & 128 \\
 & PatchRank \citep{patchrank} & $26.2 \pm 0.0$ & $53.1 \pm 0.0$ & 128 \\
 & PACT \citep{pact} & $21.8 \pm 0.0$ & $50.7 \pm 0.0$ & 128 \\
 & FiCoCo \citep{ficoco} & $24.2 \pm 0.0$ & $52.4 \pm 0.0$ & 128 \\
 & MARS (CLS-attn.\ variant) & $25.2 \pm 0.0$ & $52.8 \pm 0.0$ & 128 \\
\midrule
\multirow{11}{*}{CelebA \citep{celeba}} & Baseline & $79.6 \pm 0.0$ & $87.3 \pm 0.0$ & 256 \\
 & Masked (no prune) & $65.2 \pm 1.2$ & $75.6 \pm 1.0$ & 256 \\
 & Semantic-prune-only & $74.9 \pm 0.2$ & $85.3 \pm 0.2$ & 128 \\
 & MARS (hybrid) & $64.7 \pm 0.5$ & $74.0 \pm 0.7$ & 128 \\
 & Text-guided \citep{sparsevlm} & $74.0 \pm 0.0$ & $84.9 \pm 0.0$ & 128 \\
 & FastV \citep{fastv} & $75.6 \pm 0.0$ & $85.6 \pm 0.0$ & 128 \\
 & EViT \citep{evit} & $76.4 \pm 0.0$ & $85.5 \pm 0.0$ & 128 \\
 & PatchRank \citep{patchrank} & $78.8 \pm 0.0$ & $86.6 \pm 0.0$ & 128 \\
 & PACT \citep{pact} & $78.0 \pm 0.0$ & $86.7 \pm 0.0$ & 128 \\
 & FiCoCo \citep{ficoco} & $71.2 \pm 0.0$ & $85.1 \pm 0.0$ & 128 \\
 & MARS (CLS-attn.\ variant) & $74.8 \pm 0.0$ & $86.0 \pm 0.0$ & 128 \\

     &  &  &  &  \\[-2.2ex]
    \bottomrule
  \end{tabular}
\end{table}

\section{Full architecture $\times$ dataset results}
\label{app:arch-full}

\begin{figure}[h]
  \centering
  \includegraphics[width=0.95\linewidth]{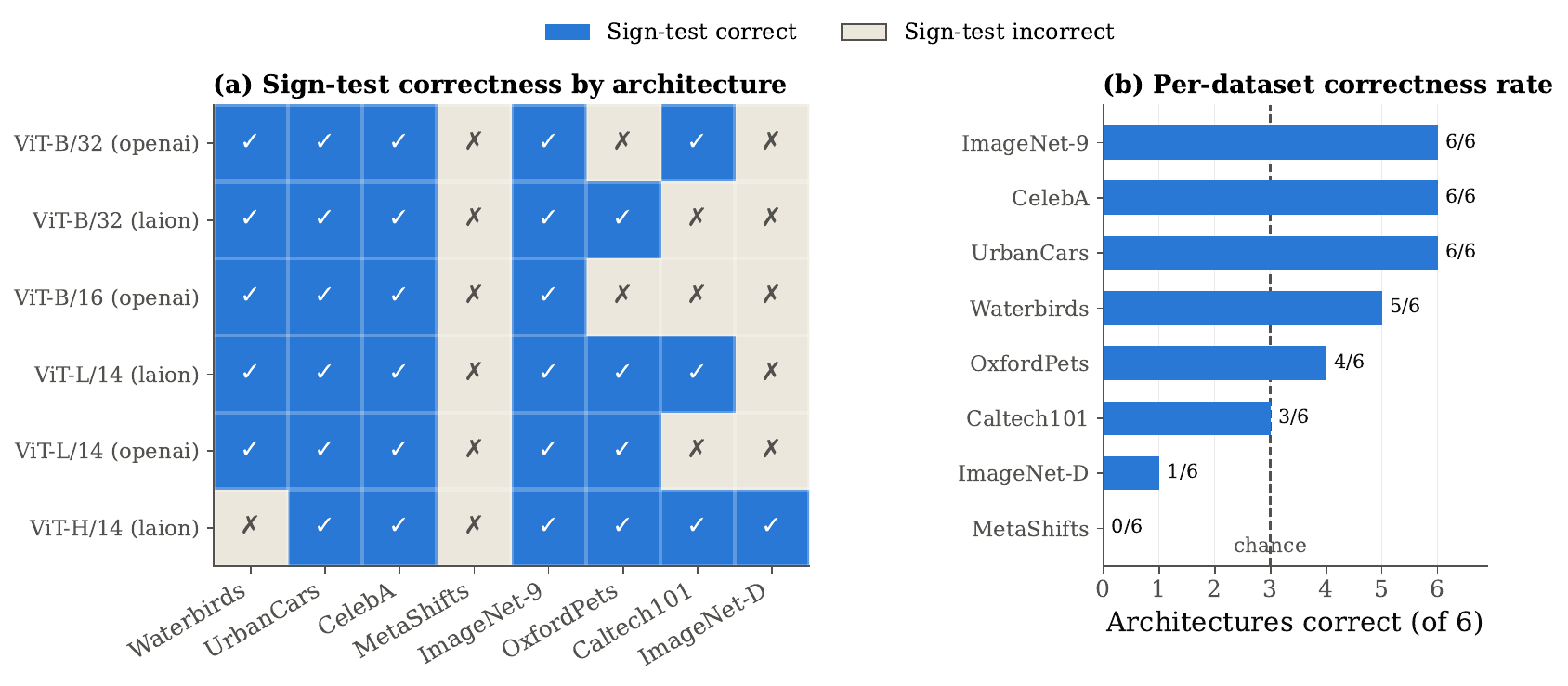}
  \caption{Sign-test correctness across 6 CLIP variants and 8 datasets (48 combinations, 31 correct, binomial $p=0.030$). \textbf{(a)} Full per-architecture breakdown: MetaShifts (column 4) is incorrect for every architecture, a reproducible structural boundary rather than architecture-specific noise. \textbf{(b)} The same data collapsed to a per-dataset correctness rate against the 3/6 chance level: three of the datasets the diagnostic was designed around sit at or near 6/6, MetaShifts sits at 0/6 (below chance), and the fine-grained classification benchmarks fall in between, where masking-based methods collapse regardless of the gate's decision.}
  \label{fig:heatmap}
\end{figure}

Table~\ref{tab:arch-full} reports the complete 48-combination result underlying Section~\ref{sec:results-arch} and Figure~\ref{fig:heatmap}: SIM, blind MARS worst-group accuracy, blind FastV worst-group accuracy, and sign-test correctness, for every (architecture, dataset) pair.

\begin{table}[h]
  \caption{Full architecture generalization results. ``Correct'' indicates the sign test ($\mathrm{SIM}>0 \Rightarrow$ MARS) selects the method with higher worst-group accuracy.}
  \label{tab:arch-full}
  \centering
  \small
  \begin{tabular}{llrrrc}
    \toprule
    Architecture & Dataset & SIM & MARS WG & FastV WG & Correct \\
    \midrule
    \multirow{8}{*}{ViT-B/32 (openai)} & Waterbirds \citep{groupdro} & +0.0287 & 63.0 & 37.0 & \checkmark \\
 & UrbanCars \citep{urbancars} & +0.0011 & 35.0 & 27.0 & \checkmark \\
 & CelebA \citep{celeba} & -0.0164 & 70.0 & 79.0 & \checkmark \\
 & MetaShifts \citep{metashifts} & +0.0089 & 56.0 & 76.0 & $\times$ \\
 & ImageNet-9 \citep{imagenet9} & -0.0039 & 50.0 & 73.0 & \checkmark \\
 & OxfordPets \citep{oxfordpets} & -0.0081 & 1.8 & 1.8 & $\times$ \\
 & Caltech101 \citep{caltech101} & -0.0050 & 0.0 & 10.5 & \checkmark \\
 & ImageNet-D \citep{imagenetd} & -0.0049 & 0.0 & 0.0 & $\times$ \\
\midrule
\multirow{8}{*}{ViT-B/32 (laion)} & Waterbirds \citep{groupdro} & +0.0287 & 53.0 & 25.0 & \checkmark \\
 & UrbanCars \citep{urbancars} & +0.0105 & 38.0 & 36.0 & \checkmark \\
 & CelebA \citep{celeba} & -0.0160 & 62.0 & 69.0 & \checkmark \\
 & MetaShifts \citep{metashifts} & +0.0138 & 21.0 & 79.0 & $\times$ \\
 & ImageNet-9 \citep{imagenet9} & -0.0119 & 40.0 & 80.0 & \checkmark \\
 & OxfordPets \citep{oxfordpets} & -0.0135 & 2.1 & 51.7 & \checkmark \\
 & Caltech101 \citep{caltech101} & -0.0149 & 0.0 & 0.0 & $\times$ \\
 & ImageNet-D \citep{imagenetd} & -0.0116 & 0.0 & 0.0 & $\times$ \\
\midrule
\multirow{8}{*}{ViT-B/16 (openai)} & Waterbirds \citep{groupdro} & +0.0315 & 35.0 & 33.0 & \checkmark \\
 & UrbanCars \citep{urbancars} & +0.0085 & 43.0 & 29.0 & \checkmark \\
 & CelebA \citep{celeba} & -0.0120 & 23.0 & 68.0 & \checkmark \\
 & MetaShifts \citep{metashifts} & +0.0102 & 60.0 & 82.0 & $\times$ \\
 & ImageNet-9 \citep{imagenet9} & -0.0115 & 35.0 & 77.0 & \checkmark \\
 & OxfordPets \citep{oxfordpets} & -0.0096 & 3.5 & 1.8 & $\times$ \\
 & Caltech101 \citep{caltech101} & -0.0155 & 0.0 & 0.0 & $\times$ \\
 & ImageNet-D \citep{imagenetd} & -0.0086 & 0.0 & 0.0 & $\times$ \\
\midrule
\multirow{8}{*}{ViT-L/14 (laion)} & Waterbirds \citep{groupdro} & +0.0443 & 59.0 & 58.0 & \checkmark \\
 & UrbanCars \citep{urbancars} & +0.0507 & 36.0 & 23.0 & \checkmark \\
 & CelebA \citep{celeba} & -0.0192 & 57.0 & 75.0 & \checkmark \\
 & MetaShifts \citep{metashifts} & +0.0250 & 71.0 & 83.0 & $\times$ \\
 & ImageNet-9 \citep{imagenet9} & -0.0230 & 30.0 & 89.0 & \checkmark \\
 & OxfordPets \citep{oxfordpets} & -0.0241 & 1.8 & 61.8 & \checkmark \\
 & Caltech101 \citep{caltech101} & -0.0195 & 0.0 & 15.8 & \checkmark \\
 & ImageNet-D \citep{imagenetd} & -0.0174 & 0.0 & 0.0 & $\times$ \\
\midrule
\multirow{8}{*}{ViT-L/14 (openai)} & Waterbirds \citep{groupdro} & +0.0488 & 59.0 & 52.0 & \checkmark \\
 & UrbanCars \citep{urbancars} & +0.0316 & 36.0 & 31.0 & \checkmark \\
 & CelebA \citep{celeba} & -0.0103 & 43.0 & 80.0 & \checkmark \\
 & MetaShifts \citep{metashifts} & +0.0121 & 49.0 & 90.0 & $\times$ \\
 & ImageNet-9 \citep{imagenet9} & -0.0096 & 30.0 & 83.0 & \checkmark \\
 & OxfordPets \citep{oxfordpets} & -0.0109 & 8.3 & 45.5 & \checkmark \\
 & Caltech101 \citep{caltech101} & -0.0050 & 5.3 & 0.0 & $\times$ \\
 & ImageNet-D \citep{imagenetd} & -0.0128 & 0.0 & 0.0 & $\times$ \\
\midrule
\multirow{8}{*}{ViT-H/14 (laion)} & Waterbirds \citep{groupdro} & +0.0847 & 12.0 & 45.0 & $\times$ \\
 & UrbanCars \citep{urbancars} & +0.0491 & 50.0 & 27.0 & \checkmark \\
 & CelebA \citep{celeba} & -0.0222 & 49.0 & 69.0 & \checkmark \\
 & MetaShifts \citep{metashifts} & +0.0454 & 60.0 & 87.0 & $\times$ \\
 & ImageNet-9 \citep{imagenet9} & -0.0357 & 20.0 & 84.0 & \checkmark \\
 & OxfordPets \citep{oxfordpets} & -0.0537 & 0.0 & 51.7 & \checkmark \\
 & Caltech101 \citep{caltech101} & -0.0244 & 0.0 & 31.6 & \checkmark \\
 & ImageNet-D \citep{imagenetd} & -0.0248 & 0.0 & 13.6 & \checkmark \\

     &  &  &  &  &  \\[-2.2ex]
    \bottomrule
  \end{tabular}
\end{table}

\section{Gated policy: visual comparison}
\label{app:gate-fig}

\begin{figure}[h]
  \centering
  \includegraphics[width=0.92\linewidth]{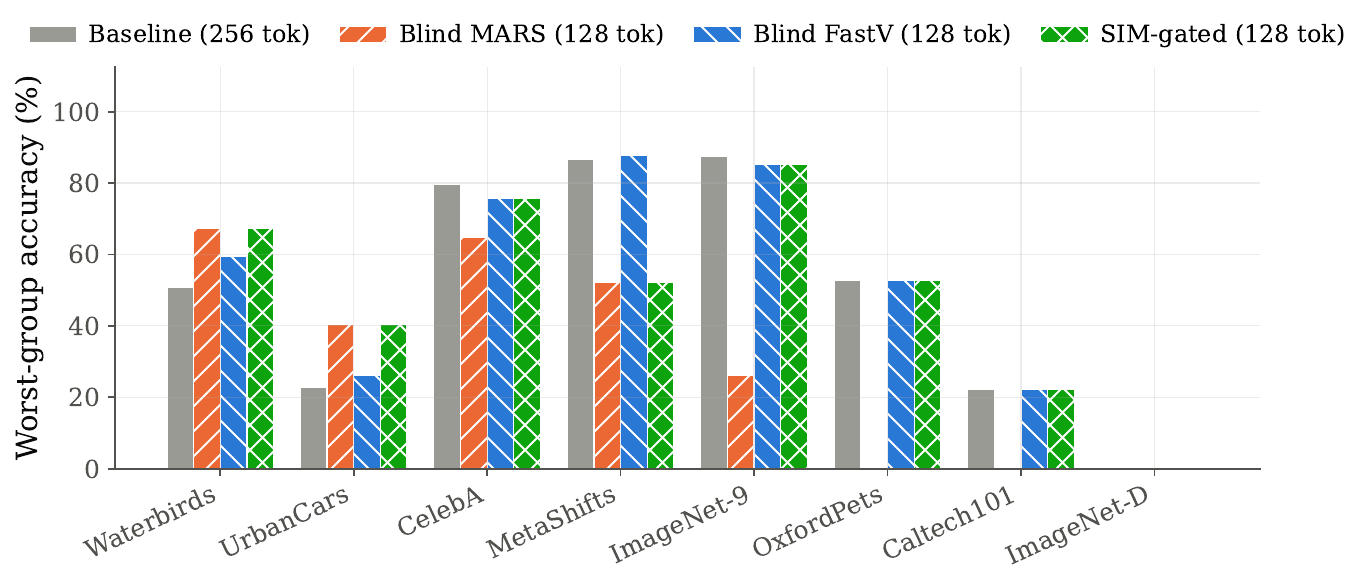}
  \caption{Worst-group accuracy under blind MARS, blind FastV, and the SIM-gated policy. The gate tracks whichever of MARS or FastV is preferable on 7 of 8 datasets; on MetaShifts (bars 4) it follows MARS, reproducing the one structural miss identified in Table~\ref{tab:sim}.}
  \label{fig:gate}
\end{figure}

Figure~\ref{fig:gate} visualizes the comparison reported in Table~\ref{tab:gate} (Section~\ref{sec:results-gate}).

\section{Ablation studies}
\label{app:ablations}

We additionally ablate three components of the MARS pipeline on Waterbirds and UrbanCars: the token-keep ratio (10--90\%), the number of $k$-means clusters ($k \in \{2,3,4,5\}$), and the Gaussian smoothing bandwidth ($\sigma \in \{0, 0.5, 0.72, 1.0, 1.5, 2.0\}$). A 33\% keep ratio modestly outperforms this sweep's own single-seed 50\%-ratio run (67.2\% vs.\ 65.6\% worst-group accuracy on Waterbirds; the main paper's 3-seed mean at 50\% is $67.3\pm2.7$, Table~\ref{tab:seeded}, consistent with this single-seed value). Neither $k=3$ nor $\sigma=0.72$, the defaults used throughout the paper, is optimal in this sweep: $k=5$ reaches 45.6\% worst-group accuracy on UrbanCars against $k=3$'s 38.4\%, and $\sigma=1.0$ modestly exceeds $\sigma=0.72$ on both datasets (68.8\% vs.\ 65.6\% on Waterbirds; 40.3\% vs.\ 38.4\% on UrbanCars). This sweep's own UrbanCars default-config run (38.4\%) sits just below the main paper's 3-seed range for the same configuration ($40.3\pm1.0$, Table~\ref{tab:seeded}); Section~\ref{sec:setup} already attributes UrbanCars' run-to-run variation to the $k$-means segmentation step's own seed sensitivity rather than resampling, which is consistent with a separately-run single-seed sweep landing just outside that range. We did not retune $k$ or $\sigma$ post hoc based on this sweep; both remain fixed at the values used throughout the rest of the paper. This shows the method is not highly sensitive to the token-keep ratio within a reasonable range, though $k$ and $\sigma$ do leave some accuracy on the table at their current defaults.

We further ablate the masking fill strategy: zeroing vs.\ mean-fill, blur-fill, and grayscale-fill. We test it on all four datasets now available to us. On the two where masking helps (Waterbirds, UrbanCars), all strategies other than grayscale-fill perform comparably. This suggests the mechanism here is the removal of background information, not the specific replacement value. On the two where masking hurts (CelebA, ImageNet-9), the result is both more consequential and more mixed. On CelebA, replacing zero-fill with grayscale-fill closes nearly all of MARS's downside risk: worst-group accuracy moves from $17.0\%$ relative below that run's own uncompressed baseline under zero-fill, to $5.9\%$ relative \emph{above} it under grayscale-fill. On ImageNet-9, the same substitution helps, but far less completely: it cuts the relative damage from $63.2\%$ below baseline to $50.4\%$ below it, still a severe failure. Fill strategy is therefore not a general fix for masking's downside risk. It substantially mitigates the risk on one of our two masking-hurts datasets, and only partially on the other. We report this as a genuine, dataset-dependent finding, complementary to, not a substitute for, the SIM-gated policy this paper's main results are built around.

\section{Full SIM predictor table}
\label{app:sim-full}

Table~\ref{tab:sim} reports the per-dataset values underlying Figure~\ref{fig:sim} (Section~\ref{sec:results-sim}).

\begin{table}[h]
  \caption{SIM and MARS's measured relative change in worst-group accuracy vs.\ the uncompressed baseline, per dataset, on OpenCLIP ViT-L/14.}
  \label{tab:sim}
  \centering
  \small
  \setlength{\tabcolsep}{4.5pt}
  \begin{tabular}{lrrccc}
    \toprule
    Dataset & SIM & Rel.\ $\Delta$WG (\%) & Predicted & Actual & Correct \\
    \midrule
    Waterbirds \citep{groupdro}  & $+0.044$ & $+25.6\%$ & help & help & \checkmark \\
    UrbanCars \citep{urbancars}  & $+0.051$ & $+82.5\%$ & help & help & \checkmark \\
    CelebA \citep{celeba}     & $-0.019$ & $-19.6\%$ & hurt & hurt & \checkmark \\
    MetaShifts \citep{metashifts} & $+0.025$ & $-39.7\%$ & help & hurt & $\times$ \\
    ImageNet-9 \citep{imagenet9} & $-0.021$ & $-70.0\%$ & hurt & hurt & \checkmark \\
    OxfordPets \citep{oxfordpets} & $-0.024$ & $-100.0\%$ & hurt & hurt & \checkmark \\
    Caltech101 \citep{caltech101} & $-0.019$ & $-100.0\%$ & hurt & hurt & \checkmark \\
    ImageNet-D \citep{imagenetd} & $-0.016$ & -$^\dagger$ & hurt & hurt & \checkmark \\
    \bottomrule
  \end{tabular}
  \\[2pt]
  \raggedright\footnotesize $^\dagger$Baseline and MARS worst-group accuracy are both 0.0\% on ImageNet-D, so relative change is undefined (0/0); classified by the raw, non-positive absolute difference instead.
\end{table}

\section{Non-stratified SIM validation}
\label{app:nonstratified}

Table~\ref{tab:nonstratified} reports the check referenced in Section~\ref{sec:limitations}: SIM computed from a flat, unstratified pool of 200 images per dataset, drawn without regard to group membership, alongside the group-stratified estimate used throughout the main results and the known MARS $\Delta$WG outcome each is being tested against. The non-stratified estimate's sign agrees with the stratified estimate on all 8 datasets, and its sign-prediction correctness (7 of 8) exactly matches the stratified version reported in Table~\ref{tab:sim}: the same single miss, MetaShifts, in both cases.

\begin{table}[h]
  \caption{Non-stratified vs.\ group-stratified SIM, 200 unstratified images per dataset. ``Correct'' indicates the non-stratified estimate's sign matches the known direction of MARS's effect on worst-group accuracy.}
  \label{tab:nonstratified}
  \centering
  \begin{tabular}{lrrrc}
    \toprule
    Dataset & SIM (stratified) & SIM (non-stratified) & Rel.\ $\Delta$WG (\%) & Correct \\
    \midrule
    Waterbirds \citep{groupdro}  & $+0.0443$ & $+0.0478$ & $+25.6\%$ & \checkmark \\
    UrbanCars \citep{urbancars}  & $+0.0513$ & $+0.0549$ & $+82.5\%$ & \checkmark \\
    CelebA \citep{celeba}     & $-0.0194$ & $-0.0246$ & $-19.6\%$ & \checkmark \\
    MetaShifts \citep{metashifts} & $+0.0249$ & $+0.0233$ & $-39.7\%$ & $\times$ \\
    ImageNet-9 \citep{imagenet9} & $-0.0213$ & $-0.0213$ & $-70.0\%$ & \checkmark \\
    OxfordPets \citep{oxfordpets} & $-0.0240$ & $-0.0243$ & $-100.0\%$ & \checkmark \\
    Caltech101 \citep{caltech101} & $-0.0188$ & $-0.0195$ & $-100.0\%$ & \checkmark \\
    ImageNet-D \citep{imagenetd} & $-0.0162$ & $-0.0162$ & -$^\dagger$ & \checkmark \\
    \bottomrule
  \end{tabular}
  \\[2pt]
  \raggedright\footnotesize $^\dagger$Baseline and MARS worst-group accuracy are both 0.0\% on ImageNet-D; relative change is undefined.
\end{table}

\section{Full per-image latency breakdown}
\label{app:latency}

Table~\ref{tab:latency} reports per-image latency for MARS's genuinely per-inference cost, referenced in Section~\ref{sec:results-efficiency}. MARS requires a full unpruned forward pass to obtain the contextualized embeddings segmentation needs, before its own pruned pass can run; the batched GPU segmentation reimplementation (Appendix~\ref{app:negative}) reduces, but does not eliminate, the resulting gap to baseline.

\begin{table}[h]
  \caption{Per-image latency (ms) on a single NVIDIA A100, OpenCLIP ViT-L/14. All rows are batch size 1 except the batched GPU segmentation row (batch 128; see Appendix~\ref{app:negative} for the full batch-size sweep).}
  \label{tab:latency}
  \centering
  \begin{tabular}{lrr}
    \toprule
    Component & Latency (ms) & Tokens \\
    \midrule
    Baseline forward pass                          & 15.7 & 256 \\
    First-pass forward for segmentation embeddings & $\approx$15.7 & 256 \\
    Segmentation step (CPU, original)               & 28.5 & - \\
    Segmentation step (batched GPU, ours, $b$=128)  & 0.81 & - \\
    MARS second-pass forward (masked, pruned)      & 11.0 & 128 \\
    FastV forward pass (single pass, no masking)   & 12.4 & 128 \\
    \midrule
    MARS total, CPU segmentation                   & 55.2 & 128 \\
    MARS total, batched GPU segmentation           & 27.5 & 128 \\
    \bottomrule
  \end{tabular}
\end{table}

\section{Removing the segmentation overhead: two unsuccessful attempts and one that succeeds}
\label{app:negative}

\textbf{GPU-resident segmentation, single-image (unsuccessful).} We reimplemented the PCA, Gaussian smoothing, and $k$-means steps to run entirely on GPU (via low-rank PCA, a depthwise-convolution Gaussian blur, and a vectorized Lloyd's-algorithm $k$-means), removing the CPU round-trip present in the original scikit-learn-based implementation. On a 48-image validation processed one image at a time (batch size 1), the GPU implementation was slower in wall-clock terms than the CPU path (161ms/image, against the 28.5ms/image CPU figure reported in Table~\ref{tab:latency}) and less faithful to it (65\% mask IoU agreement, 58.3\% vs.\ 41.7\% resulting worst-group accuracy). At the time, we attributed the slowdown to kernel-launch overhead dominating at this problem size (256 points, 3 clusters) and flagged batching, untested at the time, as the most direct plausible mitigation. The following paragraph reports that follow-up experiment.

\textbf{Batched, synchronization-free GPU segmentation (successful).} Profiling the single-image attempt above more closely than we had for the result reported a paragraph above, we found the actual bottleneck was more specific than generic kernel-launch overhead: its $k$-means initialization and convergence check called \texttt{.item()} and \texttt{.cpu()} on every iteration, each forcing a host-device synchronization that idles the GPU and hands control back to Python once per image. We rewrote $k$-means initialization and the Lloyd iteration to use only batched, GPU-resident tensor operations (\texttt{torch.multinomial} for $k$-means++ sampling across the whole batch at once, and a fixed iteration count in place of a per-iteration \texttt{torch.allclose} convergence check), and batched PCA fitting across images via \texttt{torch.pca\_lowrank}'s native batch support, so that an entire batch of images is segmented in one sequence of vectorized calls rather than one Python-level call per image.

Table~\ref{tab:batched-seg-latency} reports the resulting per-image segmentation cost as a function of batch size. It also reports a variant that reuses one fixed PCA basis, fit once on a calibration sample per dataset, instead of fitting PCA per image. Table~\ref{tab:batched-seg-accuracy} reports the correctness validation we ran before trusting either number, on 300 held-out images each from Waterbirds, UrbanCars, and CelebA. Worst-group accuracy under the per-image-adaptive batched variant matched or exceeded the CPU reference on all three datasets. Its mask IoU sits in the same 0.6--0.68 range as the original, unsuccessful single-image GPU attempt's 65\%. This tells us the batched rewrite inherits the same known approximation gap against the sklearn reference, rather than introducing a new one, and that this level of patch-level disagreement does not cost worst-group accuracy. The shared-PCA-basis variant is faster still, but its accuracy is less consistent: it varies from $-9.4\%$ to $+21.8\%$ relative across the three datasets, against the per-image variant's $0.0\%$ to $+6.5\%$. We therefore do not report it as a validated result, and flag it instead as a promising direction that needs further validation. We also note that the shared-basis variant's batch-128 measurement showed a standard deviation exceeding its own mean, an intermittent instability. Diagnosing and resolving it is a natural next step for future work; for now, we treat this specific batch-size number as an open question rather than a result to trust.

\begin{table}[h]
  \caption{Per-image segmentation cost (ms) by batch size, batched GPU reimplementation, OpenCLIP ViT-L/14 patch embeddings on real Waterbirds images (CUDA-event timing, 10 warmup + 30 measured runs). Reference: CPU (sklearn) path, 28.5ms/image at batch size 1 (Table~\ref{tab:latency}); original unbatched single-image GPU attempt, 161ms/image.}
  \label{tab:batched-seg-latency}
  \centering
  \begin{tabular}{lrr}
    \toprule
    Batch size & Per-image-adaptive PCA & Shared PCA basis \\
    \midrule
    1   & 9.06 & 7.65 \\
    8   & 1.85 & 0.95 \\
    32  & 0.98 & 0.25 \\
    64  & 0.87 & 0.14 \\
    128 & 0.81 & unreliable$^\dagger$ \\
    \bottomrule
  \end{tabular}
  \\[2pt]
  \raggedright\footnotesize $^\dagger$Measured mean was 0.27ms, but with a standard deviation (104ms) far exceeding the mean, indicating an intermittent instability at this specific batch size; we leave diagnosing it to future work (see text).
\end{table}

\begin{table}[h]
  \caption{Correctness validation for the batched GPU segmentation, 300 held-out images per dataset, against the CPU (sklearn) reference. Rel.\ $\Delta$WG is the batched variant's worst-group accuracy relative to the CPU reference's, on the same images.}
  \label{tab:batched-seg-accuracy}
  \centering
  \begin{tabular}{lrrrr}
    \toprule
    Dataset & \multicolumn{2}{c}{Per-image-adaptive PCA} & \multicolumn{2}{c}{Shared PCA basis} \\
     & Mask IoU & Rel.\ $\Delta$WG (\%) & Mask IoU & Rel.\ $\Delta$WG (\%) \\
    \midrule
    Waterbirds \citep{groupdro} & 0.680 & $+6.5\%$ & 0.666 & $+21.8\%$ \\
    UrbanCars \citep{urbancars} & 0.596 & $\phantom{+}0.0\%$ & 0.581 & $-9.4\%$ \\
    CelebA \citep{celeba}    & 0.626 & $+2.3\%$ & 0.633 & $+4.5\%$ \\
    \bottomrule
  \end{tabular}
\end{table}

With the per-image-adaptive batched variant, MARS's total per-image cost falls from 55.2ms (3.5$\times$ baseline, using the CPU segmentation step) to $\approx$27.5ms (1.75$\times$ baseline), as reported in Section~\ref{sec:results-efficiency} and Table~\ref{tab:latency}.

\textbf{Attention-based diagnostic.} We tested whether CLS attention, already computed within a single required forward pass, could replace the segmentation step entirely as the basis for a SIM-like diagnostic, using two candidate formulations: (i) the difference in text-similarity between low- and high-attention patches, and (ii) the raw correlation between per-patch attention weight and per-patch text-similarity. Evaluated against the known ground-truth direction on 5 datasets, both formulations returned a constant sign across every dataset (uniformly positive for (i), uniformly negative for (ii)), indicating neither captures dataset-specific spurious structure; both are more consistent with a generic property of CLIP's attention (plausibly related to documented ``attention sink'' effects in vision transformers) than with the phenomenon SIM is designed to measure.


\end{document}